\documentclass{ifacconf}
\usepackage{amsmath,amssymb,amsfonts}
\usepackage{algorithmic}
\usepackage{graphicx}
\usepackage{textcomp}
\usepackage{xcolor}
\usepackage{graphicx}      % include this line if your document contains figures
\usepackage{natbib}        % required for bibliography
\newcommand{\refig}[1] {Fig.~\ref{#1}}

\newcommand{\refsec}[1]{Section~\ref{#1}}

\newcommand{\loffstate}{{\mathbf q}_l}
\newcommand{\lofffp}{{\mathbf q}_l^*}
\newcommand{\linvel}{{V}}
\newcommand{\CoT}{{C}}

\newcommand{\fpstate}{{\mathbf q}_e}
\newcommand{\params}{{\mathbf p}}

\newcommand{\retmap}{{\mathbf{\bar{G}}}}
\newcommand{\stableset}[1]{\Gamma_{#1}^s}
\newcommand{\paramset}[1]{{\mathbf S}^s_{#1}}

\begin{document}
\begin{frontmatter}

\title{A compliant ankle-actuated compass walker with triggering timing control} 

%\thanks[footnoteinfo]{Sponsor and financial support acknowledgment
%goes here. Paper titles should be written in uppercase and lowercase
%letters, not all uppercase.}

\author[First]{Deniz Kerimoglu} 
\author[Second]{Ismail Uyanik}

\address[First]{School of Physics, 
   Georgia Institute of Technology, Atlanta, USA(e-mail: dkerimoglu6@gatech.edu).}
\address[Second]{Electrical and Electronics Engineering, 
   Hacettepe University, Ankara, Türkiye (e-mail: uyanik@ee.hacettepe.edu.tr)}

\begin{abstract}         Passive dynamic walkers are widely adopted as a mathematical model to represent biped walking. The stable locomotion of these models is limited to tilted surfaces, requiring gravitational energy. Various techniques, such as actuation through the ankle and hip joints, have been proposed to extend the applicability of these models to level ground and rough terrain with improved locomotion efficiency. However, most of these techniques rely on impulsive energy injection schemes and torsional springs, which are quite challenging to implement in a physical platform. Here, a new model is proposed, named triggering controlled ankle actuated compass gait (TC-AACG), which allows non-instantaneous compliant ankle pushoff. The proposed technique can be implemented in physical platforms via series elastic actuators (SEAs). Our systematic examination shows that the proposed approach extends the locomotion capabilities of a biped model compared to impulsive ankle pushoff approach. We provide extensive simulation analysis investigating the locomotion speed, mechanical cost of transport, and basin of attraction of the proposed model. 
\end{abstract}


\end{frontmatter}
%===============================================================================
\section{Introduction}
A common approach in analyzing legged locomotion is to utilize simple yet descriptive models to represent the complex dynamics of locomotion. The passive compass gait (PCG) model has been used to capture the characteristics of biped locomotion successfully \citep{Collins2005a,McGeer1990,Goswami1998}. The PCG model exhibits asymptotically stable walking behavior over slightly tilted ground. Despite its simple nature, the PCG model generates ``human-like'' walking behavior, enabling a viable basis for the analysis of biped locomotion. There are two major drawbacks of using PCG models: (1) they suffer from the narrow basin of attraction of initial conditions, and (2) they require a sloped ground to compensate for energy losses at each ground contact. Various attempts have addressed these issues by utilizing additional limbs such as torso, knees, and feet \citep{wisse2004passive,Collins2017} or using spring-loaded inverted pendulum (SLIP) legs \citep{garofalo2012walking}. These extended models can produce more anthropomorphic locomotion patterns, but they come with increased complexity, unrealistic ground contact, and a high cost of transport.

% Ankle Actuated CG Walkers Studies
Motivated by bio-mechanical observations of human walking, ankle joint actuation in PCG models provides energetically efficient locomotion as compared to other means of actuation \citep{Kuo2002,Zelik2014}. Some of these models benefit from an impulsive force applied to support the rear toe immediately before the ground collision \citep{Kuo2002,Iida2009}. As an alternative to impulsive ankle pushoff, finite-time ankle pushoff techniques have been introduced to achieve stable walking \citep{Zelik2014}. These approaches promise more natural locomotion behavior by allowing non-instantaneous pre-collision ankle pushoff \citep{Srinivasan2010,Zelik2014,Seyfarth2014}. 

These simplified ankle-actuation models are valuable for studying human muscle-tendon properties and developing more effective bipedal locomotion gaits \citep{katwal2024estimating}. Recently, \citep{dai2024multi} employed an ankle push-off method in a reduced-order model and implemented it on a large-scale Cassie robot, demonstrating that this approach enables the robot to walk at higher speeds. However, these studies overlook the passive dynamics of motion, thereby limiting the benefits of the system's inherent dynamics. Further, these studies are based on empirical observations rather than comprehensive theoretical and parametric analyses, providing limited insight into the influence of structural parameters on locomotion characteristics.

Nonetheless, there is a growing demand for practical, high-speed, and robust walking platforms \citep{siviy2023opportunities}. This requires novel approaches to design and build actuation schemes that benefit from the energy efficiency principles of underactuation and passive dynamics. Simpler mathematical models allow extensive analysis of the actuation schemes, thereby facilitating the practice of underactuation and passive dynamics. In our paper, building on our initial models \citep{KerimogluTIMC2017,kerimoglu2021efficient}, we investigate the effects of triggering timing and structural parameters of a SEA-based ankle joint on the locomotion performance of a biped model. To achieve this, we propose a new underactuated biped model inheriting the passive dynamics of the simple PCG model. The model includes an ankle joint incorporating a linear extending spring to mimic SEA. This joint allows time control of spring triggering of the SEA \citep{Pratt2004}, resembling the ankle joint in humans. We investigate gait stability as a function of spring pre-compression, triggering timing, and ankle spring stiffness. Finally, we analyze the mechanical cost of transport, basin of attraction, and locomotion speed of our model. The key contribution of this paper is the comprehensive parametric analysis of pre-compression, triggering timing, and ankle spring stiffness, providing valuable insights for the development of SEA-based bipedal robots.

\section{Triggering Controlled Ankle Actuated Compass Gait (TC-AACG) Model}
\label{sec:AACG_model}
\begin{figure}[t!]
	\centering
	\includegraphics[width=.9\columnwidth]{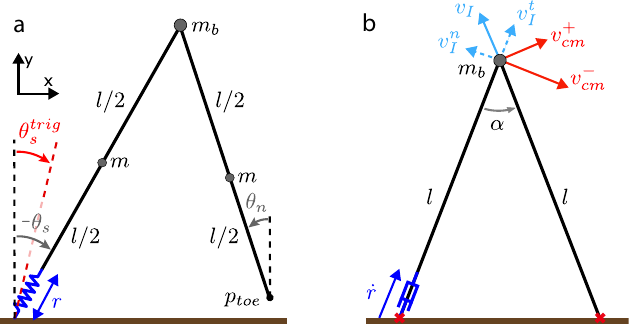}
	\caption{(a) The triggering controlled ankle actuated compass gait (TC-AACG) model with associated system parameters. (b) Configuration of the TC-AACG model at collision with associated velocity vectors.}
	\label{fig:aacg_model}
\end{figure}

\begin{figure}[h!]
	\centering
	\includegraphics[width=1\columnwidth]{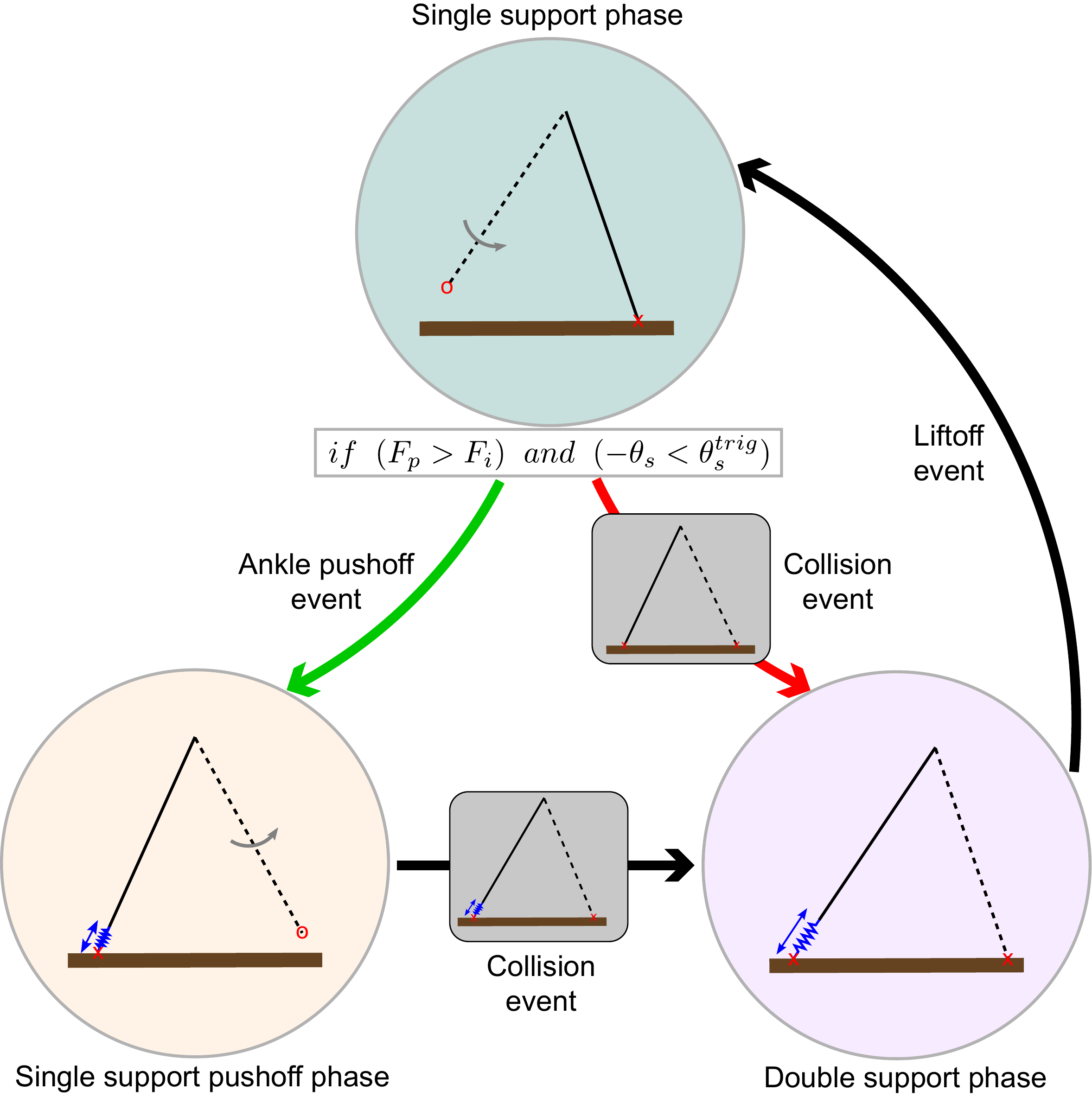}
	\caption{Locomotion phases and the transition events of the TC-AACG model. Transition from single support phase to single support pushoff or double support phase has a state-dependent nature.}
	\label{fig:aacg_phases_poff}
\end{figure}

The TC-AACG model consists of two legs attached to a point mass $m_b$, representing the torso. The legs are of length $l$, and their infinitesimal masses, $m$, are assumed to be concentrated in their center of mass (CoM). The model includes serially attached linear ankle springs with stiffness $k$ and rest length $r_0$ as illustrated in \refig{fig:aacg_model}a. During steady-state walking, the model alternates between three phases of locomotion: ``single support phase'', ``double support phase'', and ``single support pushoff phase''. The transitions from and to these phases are dictated by ``ankle pushoff'', ``collision'', and ``liftoff'' events. A detailed transition diagram for the locomotion phases is illustrated in \refig{fig:aacg_phases_poff}.
\subsection{Single Support Phase Dynamics}
\label{sec:EOMSS}
Our model exhibits PCG model dynamics in the single support phase. The parametrization of the configuration space is represented by using the vertical angles of the support and non-support legs as $\boldsymbol{q}_{ss} := [\theta_s, \theta_n]^T\;$. The detailed derivations of equations of motion for this phase can be found in \citep{Goswami1998}. 

\subsection{Single Support Pushoff Phase Dynamics}
\label{sec:SSP}
We extend the configuration space to $\boldsymbol{q}_p := [\theta_{s}, \theta_{n}, r]^T\;,$where the parameter $r$ represents the spring length. The equations of motion for this phase are obtained using the Lagrangian dynamics as
\begin{equation}
	\label{eq:ssp_dynamics}
	{{\mathbf M}_p({\mathbf q}_p)\ddot{{\mathbf q}}_p}+{{\mathbf B}_p({\mathbf q}_p,\dot{{\mathbf q}}_p)}\dot{{\mathbf q}}_p+{{\mathbf G}_{p}({\mathbf q}_{p})}={\mathbf 0}\;.
\end{equation}
Here, ${{\mathbf M}_p({\mathbf q}_p)}$ is the $3 \times 3$ mass matrix, ${{\mathbf B}_p({\mathbf q}_p, \dot{{\mathbf q}}_p)}$ is the $3 \times 3$ Coriolis matrix and ${{\mathbf G}_{p}({\mathbf q}_{p})}$ is the $3 \times 1$ vector of gravitational forces. 
\subsection{Dynamics of the Double Support Phase}
\label{sec:EOMDS}
In this phase, the model has only one degree of freedom with the configuration space defined as ${\mathbf q}_{ds} :=r$. The remaining joints depend on $r$ by the closed kinematic chain of the legs (see \refig{fig:aacg_model}). Equations of motion for this phase are obtained as
\begin{equation}
	\label{eq:collision_dynamics_appendix_GeneralForm_main}
	{{\mathbf M}_{ds}({\mathbf q}_{ds})\ddot{{\mathbf q}}_{ds}}+{{\mathbf B}_{ds}({\mathbf q}_{ds},\dot{{\mathbf q}}_{ds})}\dot{{\mathbf q}}_{ds}+{{\mathbf G}_{ds}({\mathbf q}_{ds})}=0.
\end{equation}
Here, ${\mathbf M}_{ds}({\mathbf q}_{ds})$, ${\mathbf B}_{ds}({\mathbf q}_{ds},\dot{{\mathbf q}}_{ds})$, and ${{\mathbf G}_{ds}({\mathbf q}_{ds})}$ are scalars corresponding to mass, Coriolis, and gravitational force terms, respectively. In this phase, the ankle spring on the support leg extends until it reaches its rest length. 
\subsection{Ankle Spring Pushoff Event}
\label{sec:poff_event}
This event triggers the transition from the single support phase to the single support pushoff phase. The unidirectional ankle spring on the support leg can be triggered anytime during the single support phase as depicted in \refig{fig:aacg_phases_poff}. Here, $\theta_s^{trig}$ represents the threshold variable for the triggering mechanism, implying that the spring is enabled for a possible pushoff only if $-\theta_s < \theta_s^{trig}$. Once this condition is met, the pushoff starts when the propulsive forces on the spring exceed the impeding forces. This condition is captured by $F_p > F_i \;,$ where $F_p = kr_0 + m_bl\dot{\theta_s}^2$ corresponds to the propulsive forces generated by the spring and the rotation of the body and $F_i = m_bg\cos{\theta_s}$ corresponds to impeding forces due to radial component of the gravity.
\subsection{Collision Event (Collision Map)}
\label{sec:TR}
This event triggers the transition from either the single support or the single support pushoff phases to the double support phase. There is a pivotal difference between the pre-collision and post-collision ankle pushoff phases. In the post-collision ankle pushoff case, the ankle spring is triggered right before the collision, where $r^- = 0$ and $\dot{r}^-=0$. This introduces a translational joint on the rear leg, allowing the ankle spring to freely extend along while the toes for both legs remain fixed on the ground. In the pre-collision ankle pushoff case, $r^-$ and $\dot{r}^-$ are determined during the single support pushoff phase. and the effect of impulsive forces can be captured by evaluating single support dynamics for an impulsive force acting on the toe of the swinging leg. See \citep{KerimogluTIMC2017} for detailed derivations. 

In the single support phase, the impulsive forces due to the collision event reinforce the extension of the triggered ankle spring in the rear leg if $0<\alpha<\pi/2$, where $\alpha$ is the inter-leg angle as defined in \refig{fig:aacg_model}b. The extension of the ankle spring right after the collision can be captured by $\dot{r}>0$ at the onset of the double support phase. \refig{fig:aacg_model}b shows how the impact collision affects the body's velocity vector. Here, $v_{cm}^-$ and $v_{cm}^+$ represent the pre-collision and post-collision velocity vectors of the CoM, respectively. In this representation, the velocity change due to impulsive forces can be represented by a virtual velocity vector, $v_{I}$, where $v_{cm}^+ = v_{cm}^- + v_{I}$. Utilizing kinematics of the model, one can obtain $v_{I} = l(-\dot{\theta}_s)\sin(\alpha)\;.$ The tangential component of $v_{I}$, labeled as $v_{I}^t$, is aligned with the rear leg and should be equal to the velocity of the ankle spring. This equality can be represented as
\begin{equation}
    \label{eq:coll_map}
    \dot{r} = \frac{l}{2}(-\dot{\theta}_s)\sin(2\alpha)\;.
\end{equation}
The clockwise convention for the joint angles suggests that $-\dot{\theta}_s$ and $\sin(2\alpha)$ are negative when $0<(\alpha)<\pi/2$ and $(-\dot{\theta}_s)<0$. The condition $(-\dot{\theta}_s)<0$ is met by the definition of the forward walking gait of the model. These suggest that when $0<(\alpha)<\pi/2$, the ankle spring velocity $\dot{r}$ will be positive, ensuring the extension of the ankle spring. Beyond this range, the impulsive forces yield $\dot{r}<0$, preventing the spring extension. 
\section{Gait Stability Based on Triggering Timing and Precompression}
\label{sec:stability_analysis}
\subsection{A Framework for Poincar\'{e} Stability Analysis}
\label{sec:framework}
Our framework starts by identifying the existence of limit cycles. We first define a co-dimension one subset of the state space called Poincar\'{e} section. The successive transversal of the system trajectories, which intersect the Poincar\'{e} section, yields the Poincar\'{e} map. We numerically check the stability of the fixed points to assess the stability of the proposed model.

The Poincar\'{e} section is chosen as the beginning of the single support phase. We define this instance as the liftoff state as ${\loffstate} := [\dot{\theta}_s, \theta_{n}, \dot{\theta}_n]^T$ to obtain a parametrization of the model states within a Poincar\'{e} section, $S_\Gamma$. Then, we define a parameter vector, ${\params} := [r_0, \theta_s^{trig}]$ and for two liftoff points ${\loffstate}[i]$ and ${\loffstate}[i+1]$, the Poincar\'{e} map, ${\retmap}_{\params}:S_\Gamma\rightarrow S_\Gamma$ is defined as
\begin{equation}
	\label{eq:return_map}
	{\loffstate}[{i+1}]={\retmap}_{\params}({\loffstate}[i])\;.
\end{equation}
Following this notation, the $j^{th}$ iteration of a Poincar\'{e} map during a periodic walking behavior with multiple strides is $\retmap_{\params}^j ({\loffstate})$. For each j-step fixed point, we have an associated parameter set, ${\params} := [r_0, \theta_s^{trig}]$. We combine these variables in a new parameter vector, $\fpstate = [\params, \lofffp]$, and define a set for all fixed points as
\begin{equation}
	\label{eq:Gamma_i}
	\Gamma_j := \left\{{\fpstate}:=\left[{\params},{\lofffp}\right]~|~{\retmap}_{\params}^j ({\lofffp})= {\lofffp}\right\}\;.
\end{equation}
Hence, this set, $\Gamma_j$, includes threshold angle for spring triggering, $\theta_s^{trig}$, and the spring precompression, $r_0$, for each fixed point. The linearization of ${\retmap}_{\params}^j$ around a fixed point, $\fpstate = [\params, \lofffp] \in \Gamma_j$, yields the following Jacobian equality:
\begin{equation}
	{\loffstate}[{i+1}] - {\lofffp} \approx D{\retmap}_{\params}^j|_{(\lofffp)} ({\loffstate}[i] - {\lofffp})\;,
\end{equation}
where $D{{\retmap}_{\params}^j}$ denotes the Jacobian matrix for ${\retmap}_{\params}^j$. The set of all stable fixed points are then defined as
\begin{equation}
	\label{eq:gamma_i_s}
	\stableset{j} := \left\{\fpstate~|~\fpstate\in\Gamma_j,\lambda_{m}(j, \fpstate)<1\right\}\;,
\end{equation}
where, $\lambda_{m}(j, \fpstate)$ is the maximum of the absolute values of the corresponding eigenvalues.
\subsection{Effects of Triggering Timing and Precompression of the Ankle Spring on the Gait Stability}
\label{sec:pushoff_time}
This section focuses on the TC-AACG model's gait stability based on the ankle spring's triggering timing and precompression. The model parameters for the simulations are chosen to be $m_b = 1~kg$, $m = \epsilon~kg$, $l = 1~m$, $k = 100~N/m$ where the leg mass is assumed to be infinitesimally small.
\begin{figure*}[t]
	\centering
	\includegraphics[width=1.65\columnwidth]{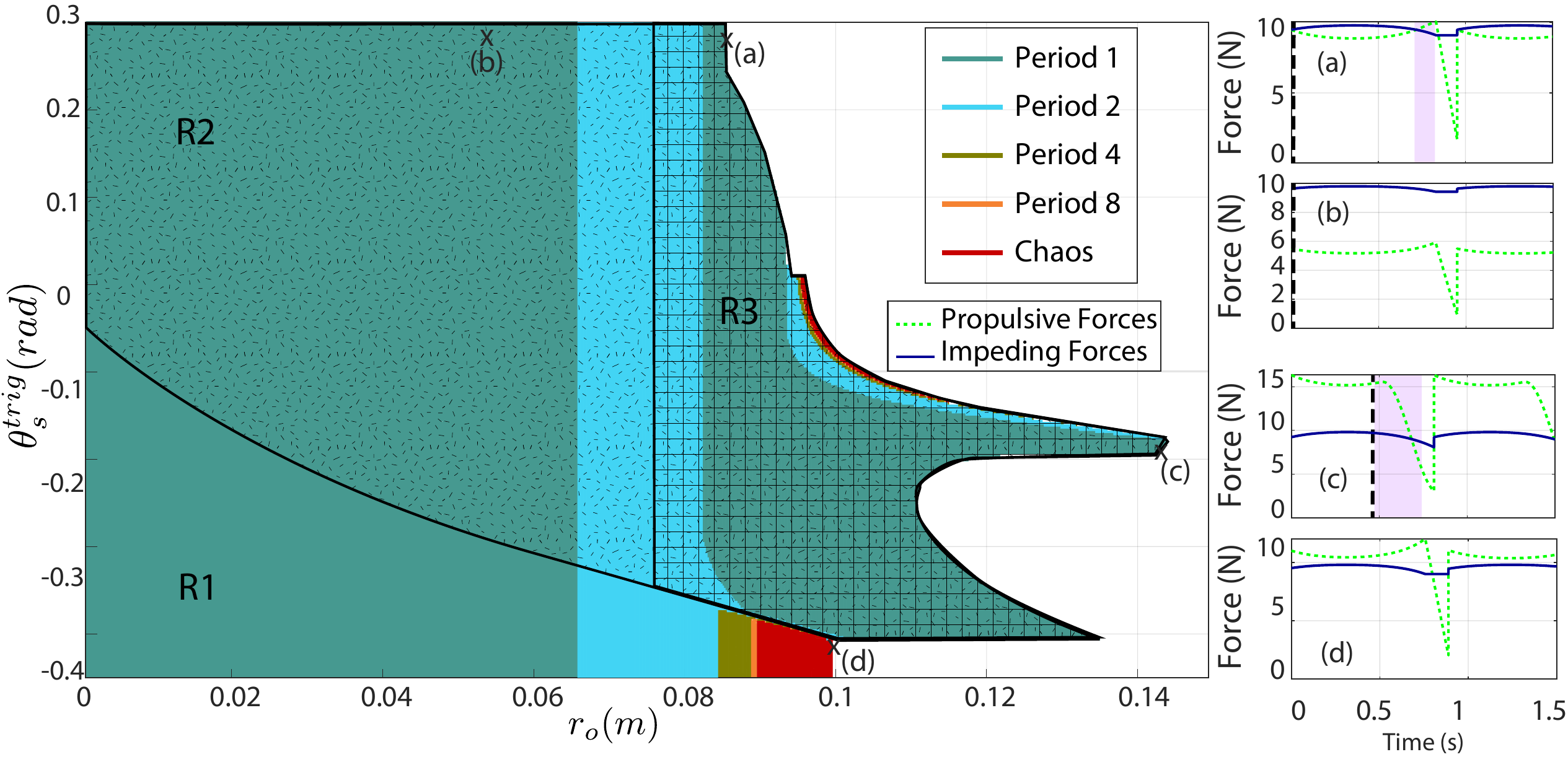}
	\caption{Fixed points of the TC-AACG model for $k=100\;N/m$ as a function of spring precompression and triggering angle along with their periodicity. $R1$, $R2$, and $R3$ represent the regions for which the spring is not triggered, spring is triggered, spring is triggered and extended in the single support phase, respectively. The subfigures illustrate the propulsive and impeding forces acting on the spring for four fixed points marked on the main figure.}
	\label{fig:fixedpt_ro_trig_m0}
\end{figure*}
To perform an illustrative analysis of the stable fixed point set $\stableset{j}$, we now define fixed point regions as 
\begin{equation}
	\label{eq:S_i}
	\paramset{j} := \left\{\params~|~\left[\params,\lofffp(\params)\right]\in{\stableset{j}}\right\}.
\end{equation}
\refig{fig:fixedpt_ro_trig_m0} illustrates $\paramset{j}$ for $j=1,2,4,8$. The teal region, $\paramset{1}$, shows the parameter values that lead to stable period-$1$ gaits. Likewise, cyan, olive, and orange regions are termed $\paramset{2}$, $\paramset{4}$, and $\paramset{8}$ and they correspond to period-$2$, period-$4$, and period-$8$ gaits, respectively. The triggering timing of the ankle spring is asserted via a threshold for the vertical angle of the support leg ($\theta_s^{trig}$). The unidirectional ankle spring is triggered once the support leg angle exceeds this threshold ($-\theta_s < \theta_s^{trig}$). However, triggering alone does not ensure the extension of the spring. The propulsive forces on the ankle spring should exceed the impeding forces to initiate the spring extension along the rear leg.

The subfigures in \refig{fig:fixedpt_ro_trig_m0} illustrate propulsive and impeding forces acting on the ankle spring as a function of time for sample fixed points, marked as $(a)$, $(b)$, $(c)$, and $(d)$. The dashed black lines represent the triggering time for the ankle spring, while the semi-transparent purple regions show the single support pushoff phase if initiated. At the fixed point $(a)$, the propulsive forces exceed the impeding forces around $t = 0.7~s$, initiating the single support pushoff phase since the spring is already triggered. In contrast, at the fixed point $(b)$, the propulsive forces cannot overcome the impeding forces. Thus, the spring does not extend until the non-support leg collides with the ground although it was triggered. At the fixed point $(c)$, the propulsive forces are greater than the impeding forces at the liftoff instant. However, spring extension is on hold until the spring is triggered at the black dashed line. As a more dramatic example, at the fixed point $(d)$, the spring is not triggered. Thus, the spring stays compressed until the non-support leg collides with the ground.

In \refig{fig:fixedpt_ro_trig_m0}, we define three regions, $R1$, $R2$, and $R3$, which illustrate the parameter sets for which the spring is not triggered, spring is triggered, spring is triggered and extended in the single support phase, respectively. Note that $R2$ and $R3$ overlap by definition since $R3$ includes the parameters for which the spring was triggered and extended.The propulsive forces on the spring can not overcome the impeding forces due to gravity until around $r_0=0.078~m$. Thus, the spring is not extended for $r_0 < 0.078~m$ even if it was triggered. With further increase of $r_0$, the propulsive forces on the ankle spring increase, injecting more energy into the model. We show that the model can sustain stable walking until $r_0 = 0.1435~m$ for $\theta_s^{trig}=-0.19~rad$. Our results suggest that the model exhibits poor locomotion performance for the fixed points when $\theta_s^{trig}>0$. These correspond to cases where the ankle pushoff is initiated before the apex point, the model's configuration where the body mass height is maximum. These ``early pushoff'' cases direct most of the spring energy to the vertical direction, elevating the body, instead of the horizontal direction. Since the cost of transport is inversely proportional to the horizontal distance traveled such cases reduce locomotion efficiency by increasing the CoT.
\begin{figure}[b!]
	\centering
	\includegraphics[width=1\columnwidth]{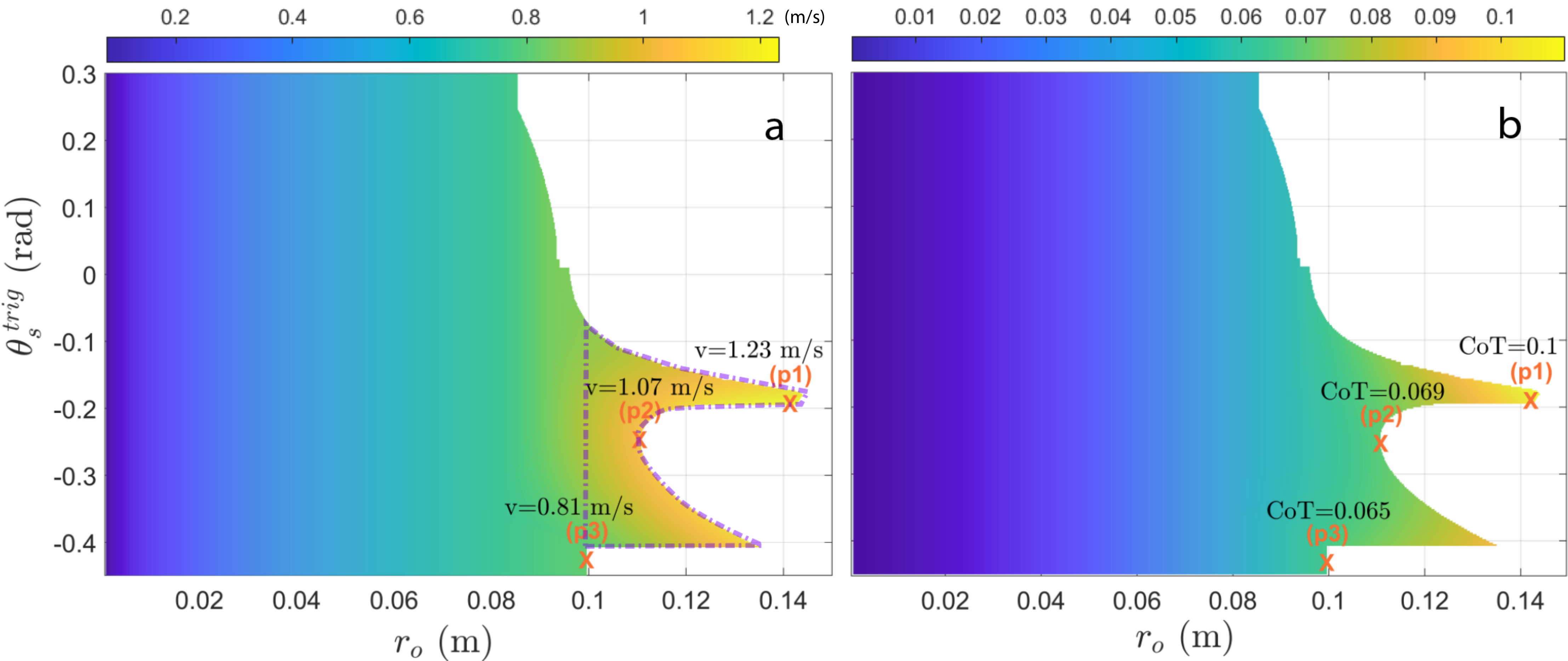}
	\caption{Horizontal speed (a) and mechanical cost of transport (b) of fixed points of the TC-AACG model as a function of spring precompression and triggering angle. The purple boundary in a) correspond to the region, $\params_{ssp}$, where the locomotion speed due to pre-collision pushoff is substantial. $p1$, $p2$, and $p3$ show the horizontal speed and mCoT for three fixed points.}
	\label{fig:fixedpt_ro_trig_vel_CoT}
\end{figure}

There is an instantaneous change in the maximum precompression around $\theta_s^{trig}=-0.4~rad$. When $\theta_s^{trig}<-0.4~rad$, the ankle spring is triggered only just before the collision, yielding a post-collision pushoff. For $\theta_s^{trig}=-0.4~rad$, the model starts experiencing pre-collision pushoff similar to impulsive pushoff \citep{Kuo2002}. The motivation for impulsive techniques was to maximize the resultant horizontal velocity via appropriate redirection of the model velocity after the collision. However, there are significant challenges in the practical implementation of impulsive techniques. Alternatively, when $\theta_s^{trig}>-0.4~rad$, there is a wide range of $r_0$ values for which the pre-collision technique we propose is applicable. Note that the proposed technique allows for generating fixed points that can utilize more spring precompression. These higher precompression regions, $\params=[{0.1<r_0<0.144,-0.4<\theta_s^{trig}<-0.1}]$, suggest compliant ankle pushoff, which starts somewhere between the apex point and collision rather than impulsive pushoff techniques. 

\begin{figure*}[h!]
	\centering
	\includegraphics[width=2\columnwidth]{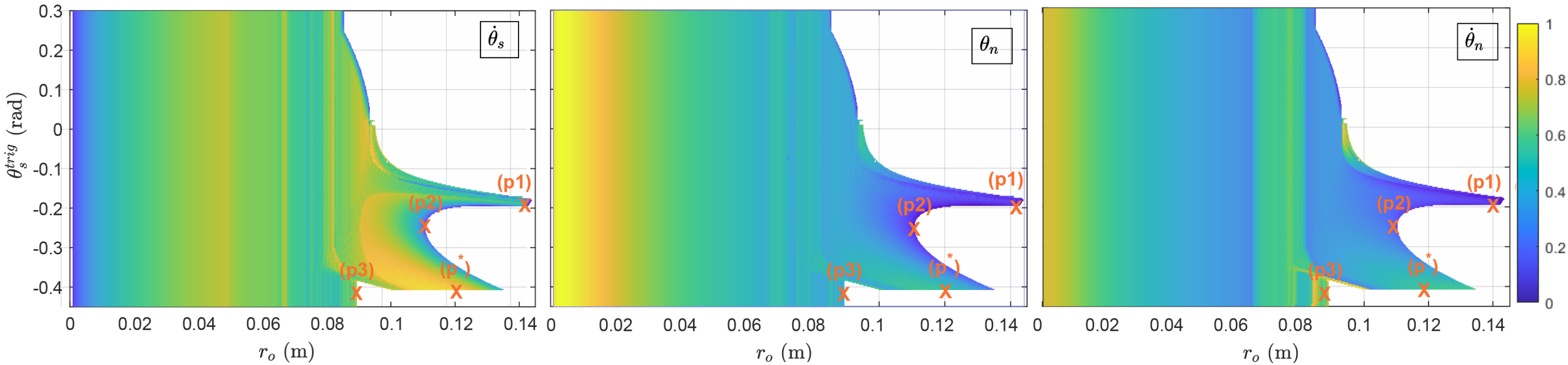}
	\caption{Area of basin of attraction of fixed points of the TC-AACG model as a function of spring precompression and triggering angle. For each subplot, $p1$, $p2$, $p3$, and $p^*$ show the area of BoA for the sample fixed points.}
	\label{fig:fixedpt_ro_trig_BoA}
\end{figure*}
\section{Efficient Locomotion via Pre-collision Ankle Pushoff}
\label{sec:Effective_locomotion}
We analyze the linear velocity, mechanical cost of transport, and basin of attraction associated with the fixed points of the model. Period-N walking traits are obtained by averaging the associated values of each stride. The linear velocity of the model for a given fixed point, $\lofffp(\params)$, is defined as $\linvel(\params) := d[i]/(t_l[i] - t_l[i-1])\;$, where $d[i]$ is the step length for the $i^{th}$ step, $t_l[i]$ and $t_l[i-1]$ are the liftoff time stamps. We obtain the linear velocity for each fixed point as shown in \refig{fig:fixedpt_ro_trig_vel_CoT} a.
Our results show that the horizontal speed of locomotion is proportional to the spring precompression. The area enclosed with purple illustrates the region where the locomotion speed due to pre-collision pushoff is substantial. In this region, $\params_{ssp}=[{ro>0.1,~-0.1>\theta_s^{trig}>-0.4}]$, we observe horizontal speeds in the range $(0.8-1.2)\;m/s$, which was not possible to obtain for $k=100\;N/m$ without pre-collision ankle pushoff (compare the green-to-yellow gradient inside the purple region with outside). The model reaches its maximum speed at $\params_1$ (see Fig. 4). Note that one should make appropriate parameter selections while considering the mechanical cost of transport for a specific parameter set. For instance, consider the speed variations along the vertical axis for $r_0 = 0.11\;m$. The same energy input yields a range of horizontal speeds based on the triggering angle selected. In this scenario, a rational choice would be $\theta_s^{trig}=-0.25~rad$, where we obtain the maximum speed for $r_0 = 0.11\;m$ (see $\params_2$). To assess the contribution of pre-collision ankle pushoff, we compare $\params_2$ with a sample point $\params_3$, which utilizes maximum spring precompression for the post-collision ankle pushoff region. The locomotion speed at $\params_2$ is significantly higher ($\approx 32\%$) than at $\params_3$.

% CoT
We examined the mechanical cost of transport (mCOT) for the same fixed points (see \refig{fig:fixedpt_ro_trig_vel_CoT}b). Combining these two figures, one can assess which precompression leads to maximum horizontal speeds with minimal cost. We first define mCoT for a given fixed point, $\lofffp(\params)$, as $\CoT(\params) = (1/2)*kr_0[i]^2/Mgd[I].$ The trade-off between the horizontal speed and the mCoT should be carefully characterized. For instance, the mCoT at $\params_2$ is significantly less than ($\approx 35\%$) the cost at $\params_1$. Selecting $\params_2$ instead of $\params_1$ drastically reduces mCoT at the cost of a $\approx 10\%$ reduction in horizontal speed. On the other hand, a comparison of $\params_2$ and $\params_3$ by looking at \refig{fig:fixedpt_ro_trig_vel_CoT} manifests the substantial contribution of the proposed pre-collision ankle pushoff. Although they have approximately identical mCoT, the model with pre-collision ankle pushoff achieves significantly higher horizontal speeds ($\approx 32\%$ more). Thus, the spring configuration of $\params_2$ triggered roughly at the midpoint between apex and collision yields the best performance for horizontal speed and mCoT.

% BoA
We also analyzed the robustness of the TC-AACG model to external disturbance in the Poincar\'{e} states by illustrating the area of basin of attraction (BoA) for $\stableset{j}$ (see \refig{fig:fixedpt_ro_trig_BoA}). For each subplot in \refig{fig:fixedpt_ro_trig_BoA}, we varied one of the coordinates of the Poincar\'{e} map on the stable limit cycle and fixed the other two. Then, the resultant areas associated with $[\dot\theta_s,~\theta_n,~\dot\theta_n]$ are normalized by multiplying them with $[5.75, 7.35, 0.195]$, respectively. Our results show that there is a large BoA for $\dot{\theta}_n$ as compared to low BoA observed for $\dot{\theta}_s$ and $\theta_n$. The increase in $r_0$, conveys different effects on the BoA of each Poincar\'{e} state. The BoA for $\dot{\theta}_s$ increases until $r_0 = 0.05\;m$ but then decreases until $r_0 = 0.078\;m$, where the model starts experiencing pre-collision ankle pushoff as depicted in R3 of \refig{fig:fixedpt_ro_trig_m0}. However, for $\theta_n$ and $\dot{\theta}_n$, the BoA decreases until $r_0 = 0.06\;m$ and remains almost fixed beyond this point. These results imply the necessity of considering a possible trade-off between the BoA of the model states when controlling the precompression of the ankle spring. Therefore, even though \refig{fig:fixedpt_ro_trig_vel_CoT} suggests that $\params_2$ yields the best performance for horizontal speed and mCoT, the associated area of BoA for all states is quite narrow. Alternatively, the spring configuration $\params^*$ triggered close to the collision offers significant robustness against external disturbances with a broader area of BoA for v$\approx1~m/s$ and mCoT$=0.08$. 

\begin{figure}[h!]
	\centering
	\includegraphics[width=1\columnwidth]{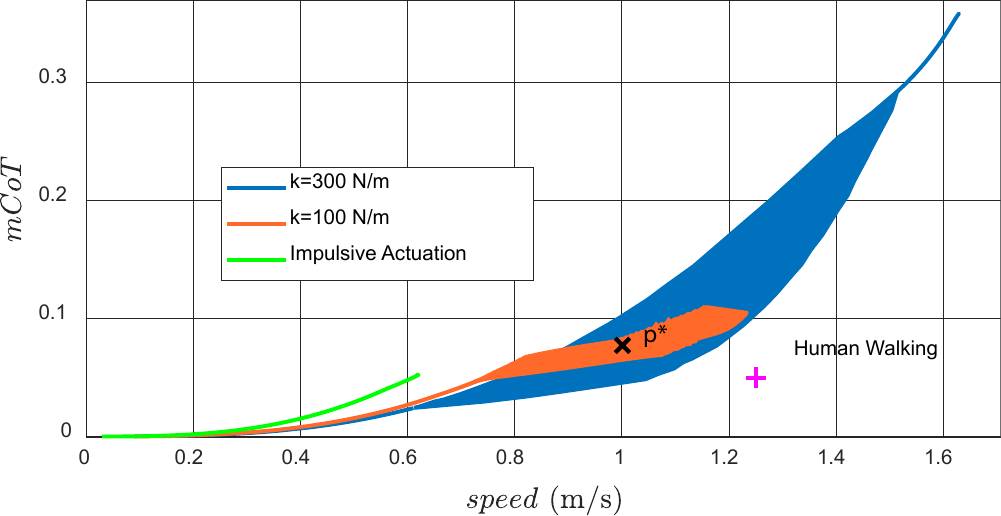}
	\caption{The locomotion performances of the proposed technique against impulsive actuation.}
	\label{fig:LocomotionPerf}
\end{figure}
To complete our analysis on the trade-off between the speed, efficiency and robustness, we illustrate the range of mCoT and horizontal speed that our model can achieve for a desired horizontal speed or mCoT in \refig{fig:LocomotionPerf} for different values of spring constant: $k=100\;N/m$, $k=300\;N/m$ and impulsive actuation. Here, $k=300\;N/m$ is a sample stiffness value for which the TC-AACG model performs well (not the optimal case). In contrast, $k=100\;N/m$ is a sample case for which the TC-AACG model can not completely benefit from compliant ankle actuation. Different from these, the green curve represents the stable fixed points of impulsive ankle pushoff technique which are obtained by evaluating the associated dynamics in \citep{Kuo2002} with the model assumptions in \refsec{sec:AACG_model}. The proposed compliant ankle actuation at $k=300\;N/m$ can obtain less mCoT for a wide range of horizontal speeds compared to impulsive pushoff, representing $k=\infty\;N/m$. Similarly, we can obtain higher horizontal speeds for a given mCoT as compared to the impulsive method. More importantly, the spring configuration $p^*$ in Fig. 5 exhibited a slight compromise on the speed and energy efficiency but suggested increased locomotion robustness. Our proposed method enables the exploration of a wide range of stable limit cycles, allowing for a trade-off between locomotion speed, efficiency, and robustness to achieve performance comparable to human locomotion.

\section{Conclusion}
\label{sec:Conclusion}
In this paper, we proposed a new bipedal walking model, triggering controlled ankle actuated compass gait (TC-AACG), which allows adjusting the pre-collision triggering timing of the ankle spring. Our approach was motivated by the need for physically realizable techniques for energy regulation in bipedal robot models. We conducted an extensive parametric analysis to examine the effects of precompression, ankle spring triggering timing, and spring stiffness on the model, exploring the trade-offs between locomotion characteristics such as horizontal speed, mechanical cost of transport, and basin of attraction. These models enabled extensive simulations to investigate locomotion stability and control principles, which would be difficult to achieve through physical experiments. 

We utilized Poincar\'{e} analysis to assess the stability of the fixed points of the model. The benefits of pre-collision ankle actuation have been shown previously via impulsive/axial ankle force or torsional ankle springs using optimization methods. However, to the best of the authors' knowledge, an extensive investigation of locomotion characteristics such as speed, efficiency, and robustness has not been previously presented. Thus, the results of this analysis will be fundamental for designing an optimal locomotion controller with these models. This research highlights key design strategies for improving locomotion efficiency. We showed that the proposed pre-collision ankle pushoff enables higher locomotion speeds for a given mechanical cost of transport as compared to impulsive in the literature \citep{Kuo2002}. To reduce the mechanical cost of transport and achieve desired locomotion speed, passivity and underactuation principles should be applied through techniques like pre-collision ankle pushoff. Initiating ankle pushoff around the midpoint between apex and collision significantly improves speed and efficiency. Moreover, we showed that slightly reducing speed and energy efficiency could lead to more robust locomotion.

\bibliography{ifacconf}             
                                                      
\end{document}